\documentclass[conference]{IEEEtran}
\IEEEoverridecommandlockouts

\usepackage{booktabs}
\usepackage{cite}
\usepackage{comment}
\usepackage{amsmath,amssymb,amsfonts}
\usepackage{algorithmic}
\usepackage{graphicx}
\usepackage{textcomp}
\usepackage{xcolor}
\usepackage{hyperref}
\usepackage{multirow}
\usepackage[linesnumbered,ruled,vlined]{algorithm2e}
\usepackage{subcaption}
\usepackage{tikz}
\def\BibTeX{{\rm B\kern-.05em{\sc i\kern-.025em b}\kern-.08em
    T\kern-.1667em\lower.7ex\hbox{E}\kern-.125emX}}
\usepackage{balance}

\hypersetup{
  colorlinks=true,
  linkcolor=blue,
  citecolor=blue,
  filecolor=magenta,
  urlcolor=blue
}

\makeatletter
\def\ps@IEEEtitlepagestyle{%
  \def\@oddfoot{\mycopyrightnotice}%
  \def\@evenfoot{}%
}
\def\mycopyrightnotice{%
  \gdef\mycopyrightnotice{}%
}
\makeatother

\usepackage{eso-pic}
\newcommand\AtPageUpperMyleft[1]{\AtPageUpperLeft{
 \put(\LenToUnit{1cm},\LenToUnit{-1cm}){ 
     \parbox{0.5\textwidth}{\raggedright\fontsize{9}{11}\selectfont #1}} 
 }}
\newcommand{\conf}[1]{
\AddToShipoutPictureBG*{
\AtPageUpperMyleft{#1}
}
}

\def\BibTeX{{\rm B\kern-.05em{\sc i\kern-.025em b}\kern-.08em
    T\kern-.1667em\lower.7ex\hbox{E}\kern-.125emX}}
\begin{document}

\title{Quantitative Currency Evaluation in Low-Resource Settings through Pattern Analysis to Assist Visually Impaired Users\\
}
\conf{This work has been submitted for possible publication. Copyright may be transferred without notice, after which this version may no longer be accessible.}

\author{\IEEEauthorblockN{Md Sultanul Islam Ovi}
\IEEEauthorblockA{\textit{Department of Computer Science} \\
\textit{George Mason University}\\
Fairfax, USA \\
\href{mailto:movi@gmu.edu}{movi@gmu.edu}}
\and
\IEEEauthorblockN{Mainul Hossain}
\IEEEauthorblockA{\textit{Department of Computer Science} \\
\textit{George Mason University}\\
Fairfax, USA \\
\href{mailto:mhossa6@gmu.edu}{mhossa6@gmu.edu}}
\and
\IEEEauthorblockN{Md Badsha Biswas}
\IEEEauthorblockA{\textit{Department of Computer Science} \\
\textit{George Mason University}\\
Fairfax, USA \\
\href{mailto:mbiswas2@gmu.edu}{mbiswas2@gmu.edu}}
}

\maketitle

\begin{abstract}
Currency recognition systems often overlook usability and authenticity assessment, especially in low-resource environments where visually impaired users and offline validation are common. While existing methods focus on denomination classification, they typically ignore physical degradation and forgery, limiting their applicability in real-world conditions. This paper presents a unified framework for currency evaluation that integrates three modules: denomination classification using lightweight CNN models, damage quantification through a novel Unified Currency Damage Index (UCDI), and counterfeit detection using feature-based template matching. The dataset consists of over 82,000 annotated images spanning clean, damaged, and counterfeit notes. Our Custom\_CNN model achieves high classification performance with low parameter count. The UCDI metric provides a continuous usability score based on binary mask loss, chromatic distortion, and structural feature loss. The counterfeit detection module demonstrates reliable identification of forged notes across varied imaging conditions. The framework supports real-time, on-device inference and addresses key deployment challenges in constrained environments. Results show that accurate, interpretable, and compact solutions can support inclusive currency evaluation in practical settings.
\end{abstract}

\begin{IEEEkeywords}
Visually Impaired Accessibility, Inclusive Financial Technology, Unified Currency Damage Index (UCDI), Real-World Currency Evaluation, Computer Vision, Deep Learning
\end{IEEEkeywords}

\section{Introduction}

Despite widespread adoption of digital payments, physical currency remains a primary medium of exchange across many regions, especially where electronic infrastructure is limited or unreliable \cite{tanzim2023}. In these settings, accurate and efficient handling of paper money is essential not only for individuals but also for institutions, retail workers, and service providers operating in cash-based environments. For people who are visually impaired, interacting with physical currency introduces additional barriers. Identifying denominations, assessing whether a note is still usable, and verifying authenticity all rely on visual cues that may be inaccessible or ambiguous. 

Assistive technologies have made incremental progress, particularly in denomination recognition. However, most systems are narrow in scope, often designed for a single task, and tend to overlook real-world challenges like torn or faded notes, uneven lighting, and low-quality image inputs. Several works have explored methods to detect tears, stains, or physical degradation in currency \cite{krishnapriya2024soiled}, but these are rarely integrated into unified tools. Limitations become more pronounced when the solutions rely on large models that are impractical for mobile or embedded deployment, making them less useful for people who need reliable offline performance on lightweight devices. 

An effective solution must answer three essential questions: What is the denomination of this note? Can it still be used in its current condition? Is it genuine? These questions are not only relevant for visually impaired users but also critical for broader applications in finance, commerce, and public services that depend on accurate and fast physical currency validation. A tool that cannot assess all three aspects will offer limited utility in daily use. 

We present a unified framework that addresses this gap by combining denomination classification, physical damage detection, and counterfeit verification in a single system. Each module is designed for low-resource environments, with a focus on high accuracy, fast inference, and deployment readiness for handheld or embedded platforms. The key innovation lies in the damage assessment component, which introduces a structured scoring system that accounts for visual degradation and structural defects.

We propose the \textbf{Unified Currency Damage Index (UCDI)}, a continuous metric that quantifies a note’s usability by combining binary mask differences, color degradation, symbolic feature loss, and spatial inconsistencies. This enables consistent and interpretable usability evaluation across a wide range of damage types, from minor stains to significant tears. \\

Our contributions include:
\begin{itemize}
    \item A lightweight convolutional architecture for denomination recognition with competitive performance on resource-constrained devices.
    \item A novel damage quantification framework that integrates structural, chromatic, and symbolic degradation into a single score (UCDI).
    \item An efficient counterfeit detection module that identifies forgeries based on fine-grained texture and visual markers.
\end{itemize}

This system is built for real-world conditions and generalizes across diverse currencies and environments. It is designed to support visually impaired individuals who rely on accessible tools for cash handling, as well as broader use cases in retail, banking, and public service domains where accurate, portable, and inclusive currency evaluation remains a priority.

\section{Literature Review}

Recent research in automated currency analysis has advanced across three primary fronts: denomination classification, counterfeit detection, and, to a lesser extent, damage assessment. Early efforts focused on maximizing classification accuracy while reducing computational overhead. Pachón et al.\ \cite{pachon} showed that a compact CNN could match the accuracy of ResNet18 for Colombian banknotes while achieving significant speedups on both CPU and GPU. Jaman et al.\ \cite{jaman2025bd} built an offline Android application that reached 98.5\% accuracy on local currency classification. These models highlight the trade-off between architectural depth and inference latency, a core challenge for mobile deployment.

A parallel line of work has focused on assistive applications for visually impaired users. Tasnim et al.\ \cite{tasnim2021bangladeshi} presented a CNN classifier that delivered both textual and auditory feedback with 92\% accuracy. Das et al.\ \cite{das2023utilizing} employed ResNet50 to achieve high precision in note identification. Neto et al.\ \cite{neto} and Awad et al.\ \cite{awad} extended mobile solutions to Brazilian and Iraqi currencies, respectively, demonstrating successful real-time classification under field conditions. Bolaños-Fernández and Bacca-Cortes \cite{bolanos} proposed CopReader, which integrates MobileNet and YOLOv5 in a mobile app with audio outputs, enhancing autonomy for users with visual impairments. These contributions illustrate the feasibility of lightweight, accessible tools, though most limit their scope to denomination recognition alone.

Counterfeit detection systems have adopted a more varied approach, blending classical and deep learning methods. Mukundan et al.\ \cite{mukundan} built a Raspberry Pi-based hyperspectral scanner to detect counterfeit Taiwanese currency using spectral features between 400 and 500 nm. Mehta and Bhalla \cite{mehta} proposed a hybrid CNN-SVM architecture to classify fake notes across five classes. Giri et al.\ \cite{giri} employed a YOLOv8-based segmentation model for Nepali Rs 1000 notes, reporting significant improvements over YOLOv5 baselines on both precision and recall. Antonius et al.\ \cite{antonius} introduced DeepCyberDetect, a composite framework combining GANs, CNNs, RNNs, and African Buffalo Optimization, achieving 99.2\% accuracy across varied currency types and conditions.

\begin{figure}[ht]
    \centering
    \includegraphics[width=\linewidth]{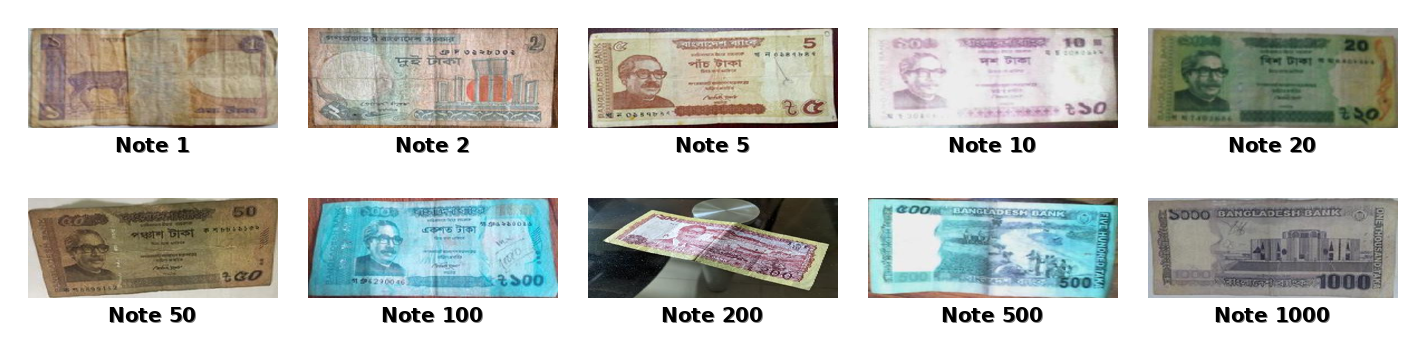}
    \caption{Representative Samples from Each Currency Class in the Standardized Combined Dataset}
    \label{fig:sample_images}
\end{figure}

Other studies have emphasized cross-currency generalization and accessibility. Pham et al.\ \cite{pham2020, pham2022} developed CNN models that generalized well across multiple currencies, requiring no specialized hardware, and supporting visually impaired users. Park et al.\ \cite{park2020} proposed a three-stage pipeline using Faster R-CNN and ResNet for improved counterfeit detection under complex imaging conditions. Bhatia et al.\ \cite{bhatia} compared KNN, SVC, and Gradient Boosting on Indian banknotes, with KNN achieving 99.9\% accuracy. Kanawade et al.\ \cite{kanawade} demonstrated that fine-tuned ResNet-50 surpassed GoogleNet and VGG-16 in counterfeit detection, confirming the advantage of deeper residual networks for capturing subtle features like microprinting.

Despite the progress in classification and forgery detection, damage assessment remains a relatively unexplored domain. Most systems assume clean, undamaged currency or treat severe damage as a classification failure. Pham et al.\ \cite{pham2018deep} addressed banknote fitness using 1D visible-light image sensors, achieving over 97\% accuracy, but their approach required specialized hardware. Meshram et al.\ \cite{meshram2023comprehensive} emphasized the lack of robust datasets for damaged Indian notes. Suardi et al.\ \cite{suardi2023bank} and Farooq et al.\ \cite{farooq2023innovating} attempted visual damage detection but achieved limited accuracy, relying on basic heuristics rather than structured degradation models.

To date, to the best of our knowledge, there is no established framework that performs multi-faceted damage analysis at the spatial, color, and symbolic levels while also integrating classification and counterfeit detection. Our proposed system fills this gap by introducing a fully automated pipeline that aligns damaged currency with clean templates and extracts structural, chromatic, and feature-based cues to compute a unified usability score. This damage index offers a consistent, interpretable metric for evaluating note condition and supports practical decision-making in visually assistive or financial screening contexts.

\section{Methodology}

Building on the limitations identified in prior work, we propose an integrated framework that performs denomination classification, damage assessment, and counterfeit detection in a unified pipeline. Unlike existing systems that treat these tasks in isolation or rely on hardware-heavy solutions, our approach emphasizes modularity, interpretability, and deployment efficiency on resource-constrained devices.

The damage assessment module introduces the \textbf{Unified Currency Damage Index (UCDI)}, a continuous usability score derived from structural, chromatic, and symbolic degradation cues. Each component of the system is designed to function independently while contributing to end-to-end analysis, enabling both targeted evaluation and full-spectrum currency screening.

\begin{table}[ht]
\centering
\caption{Summary of Datasets Used for Currency Classification, Including Total Samples, Number of Classes, and Image Resolutions}
\begin{tabular}{|c|c|c|c|}
\hline
\textbf{Dataset}       & \textbf{Total Images} & \textbf{Classes} & \textbf{Image Ratio} \\ \hline
Dataset 01             & 1,970                 & 9                & 120 x 250            \\ \hline
Dataset 02             & 10,000                & 10               & 224 x 224            \\ \hline
Dataset 03             & 70,542                & 8                & 200 x 250            \\ \hline
\hline
Combined Dataset       & 82,512                & 10               & 224 x 224            \\ \hline
\end{tabular}
\label{tab:datasets_overview}
\end{table}

\subsection{Dataset Overview and Preprocessing}

Three publicly available datasets were used to train and evaluate the classification module. Table~\ref{tab:datasets_overview} summarizes the dataset properties, and Figure~\ref{fig:sample_images} shows representative standardized samples. \textbf{Dataset 1} \cite{nsojib2023bangla} introduces real-world variability through mobile-captured images but does not include denomination 200. \textbf{Dataset 2} \cite{mohammed2022augmented} offers a balanced, high-quality collection across all denominations. \textbf{Dataset 3} \cite{tasnim2023bangladeshi} provides intra-class diversity with both old and redesigned versions but excludes denominations 1 and 200. All three were resized and merged into a 10-class unified dataset of 82,512 images used for classification.

\begin{figure}[htbp]
    \centering
    \includegraphics[width=\linewidth]{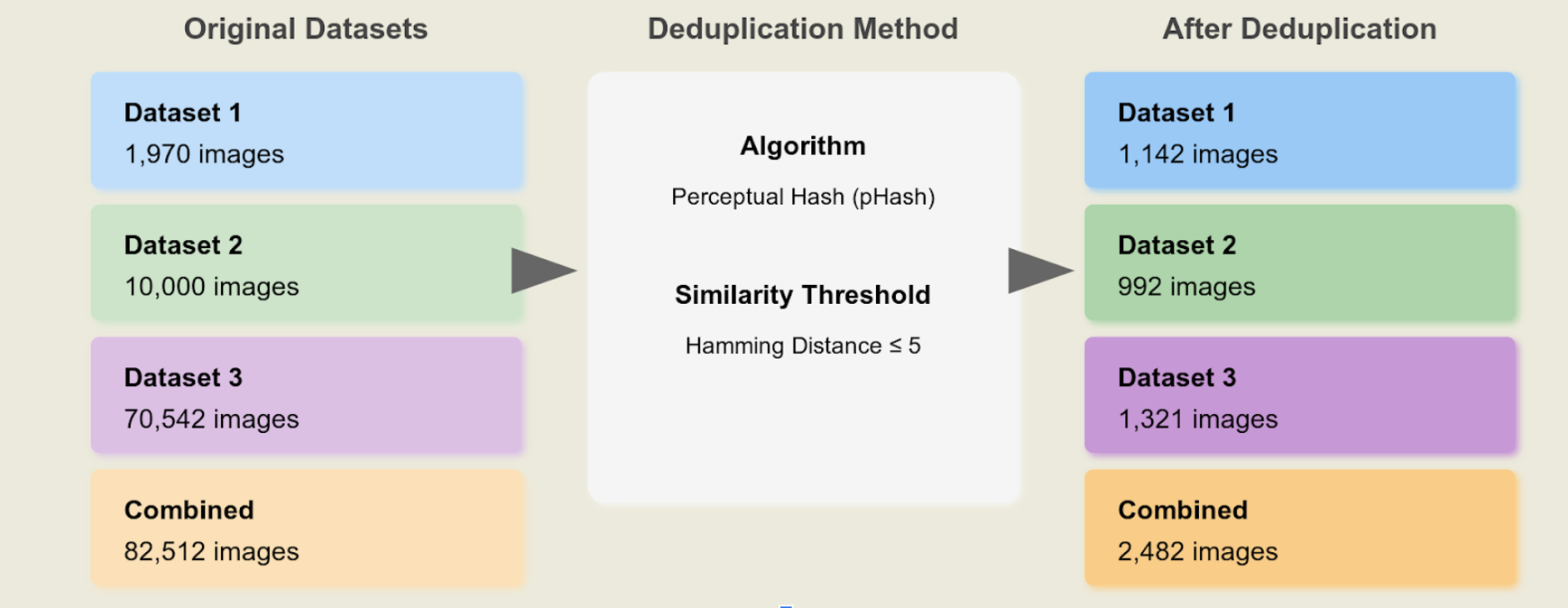}
    \caption{Overview of the Image Deduplication Pipeline Implemented Across All Datasets}
    \label{fig:dedup_overview}
\end{figure}

\textbf{1. Dataset Analysis:} The source datasets contained visually similar or identical images, introducing potential redundancy and risk of data leakage. To address this, a perceptual hash-based deduplication pipeline was applied. Figure~\ref{fig:dedup_overview} shows the overall process. The method uses perceptual hashes with a Hamming distance threshold of 5 to detect duplicates; this value was empirically selected to balance between over-pruning and missed near-duplicates. A class-wise hash dictionary is maintained to retain only distinct entries. The full procedure is described in Algorithm~\ref{alg:dedup}, which was implemented with parallel processing for scalability. After individual deduplication, a cross-dataset check was conducted to remove overlaps between sources. This ensured strict separation of training, validation, and test sets, preserving the reliability of model evaluation.

\begin{algorithm}[htbp]
\SetAlgoLined
\SetKwInOut{Input}{Input}
\SetKwInOut{Output}{Output}
\Input{Source dataset $D$, similarity threshold $t$}
\Output{Deduplicated dataset $D'$}
$D' \gets \emptyset$\; \tcp{Initialize deduplicated dataset}
\ForEach{class $c$ in $D$}{
    $H \gets \emptyset$\; \tcp{Initialize hash dictionary for class $c$}
    \ForEach{image $i$ in class $c$}{
        $h \gets \text{CalculatePerceptualHash}(i)$\;
        $similar \gets \text{FindSimilarHash}(H, h, t)$\;
        \If{$similar = \text{None}$}{
            $H[h] \gets i$\; \tcp{Store hash and corresponding image}
            $D' \gets D' \cup \{i\}$\; \tcp{Add image $i$ to deduplicated dataset}
        }
    }
}
\Return{$D'$}\; \tcp{Return the deduplicated dataset}
\caption{Image Deduplication Process}
\label{alg:dedup}
\end{algorithm}

\textbf{2. Dataset Splitting:}  
The combined dataset was divided into training, validation, and test sets in an 80:10:10 ratio using stratified sampling to preserve class distribution.

\textbf{3. Image Enhancement:}  
Each image underwent a sequence of operations to reduce noise and improve visual clarity:

\begin{itemize}
   \item \textbf{Median Blur:} Suppressed impulsive noise while preserving edges. For kernel size $k = 3$, each pixel was updated using:
   \begin{equation}
   \begin{aligned}
   I'(x,y) = \text{median}\{I(x+i,y+j) \mid \\
   i,j \in [-\tfrac{k-1}{2},\tfrac{k-1}{2}]\}
   \end{aligned}
   \end{equation}

   \item \textbf{Sharpening:} Applied a fixed kernel $K$ to enhance local details:
   \begin{equation}
   I'(x,y) = \sum_{i=-1}^{1}\sum_{j=-1}^{1} K(i,j) I(x-i,y-j)
   \end{equation}
   where $K = \begin{bmatrix} -1 & -1 & -1 \\ -1 & 9 & -1 \\ -1 & -1 & -1 \end{bmatrix}$.

   \item \textbf{Contrast Stretching:} Performed on each RGB channel using:
   \begin{equation}
   I'_c(x,y) = 255 \times \frac{I_c(x,y) - p_2}{p_{98} - p_2}
   \end{equation}
   with $p_2$ and $p_{98}$ as the 2nd and 98th percentiles.

   \item \textbf{CLAHE:} Applied in LAB color space with clip limit 2.0 and tile size 8×8. The enhanced output is computed by:
   \begin{equation}
   I'(x,y) = \text{CDF}(I(x,y)) \cdot (I_{\text{max}} - I_{\text{min}}) + I_{\text{min}}
   \end{equation}
   and the clipped histogram:
   \begin{equation}
   h'(i) = \min(h(i), cl) + \frac{\sum_{j=0}^{L-1} \max(h(j) - cl, 0)}{L}
   \end{equation}
   where $cl$ is the clip limit and $L$ is the number of bins.
\end{itemize}

\textbf{4. Data Augmentation:}  
To improve generalization, each training image was augmented using the following stochastic transformations:
\begin{itemize}
   \item Random resized crop (scale 0.8 to 1.0)
   \item Random rotation (±15 degrees)
   \item Horizontal flip
   \item Color jitter (brightness, contrast, saturation: 0.2; hue: 0.1)
   \item Affine transform (translation $\leq$10\%, scale 0.9 - 1.1, shear $\leq$10°)
   \item Random erasing (50\% probability, 2-15\% area)
\end{itemize}

All images were finally resized to 224×224 pixels and normalized using ImageNet statistics to ensure compatibility with pre-trained convolutional backbones.

\subsection{Classification Framework}

\begin{figure}[htbp]
    \centering
    \includegraphics[width=0.9\linewidth]{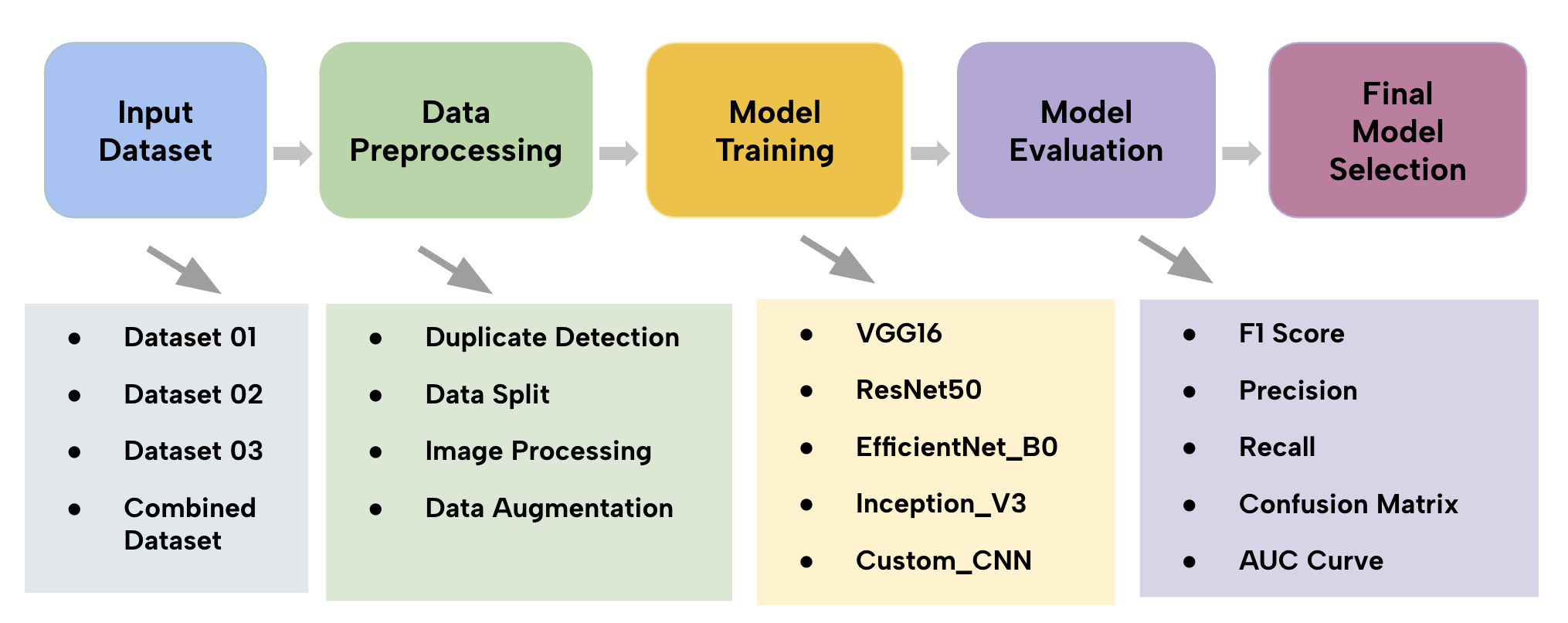}
    \caption{End-to-end currency classification workflow, including dataset curation, preprocessing, model training, evaluation, and final model selection.}
    \label{fig:classification_pipeline}
\end{figure}

After preprocessing, a multi-model classification pipeline was developed to recognize currency denominations. As shown in Figure~\ref{fig:classification_pipeline}, the pipeline processes standardized images to produce denomination labels.

\textbf{1. Deep Learning Models.}  
We fine-tuned four convolutional models pretrained on ImageNet for multi-class currency classification: VGG16~\cite{simonyan2014very}, ResNet50~\cite{he2016deep}, InceptionV3~\cite{szegedy2016rethinking}, and EfficientNetB0~\cite{tan2019efficientnet}. VGG16 uses stacked 3×3 convolutions and max pooling, with approximately 138 million parameters. ResNet50, with 25 million parameters, includes residual connections to improve gradient flow in deeper networks. InceptionV3 captures multi-scale features through parallel convolutional branches and has 23 million parameters. EfficientNetB0 applies compound scaling to balance accuracy and efficiency, using only 5.3 million parameters. For all models, we replaced the final classification layers and fine-tuned the upper blocks on the merged dataset.

\begin{figure}[htbp]
   \centering
   \includegraphics[width=0.9\linewidth]{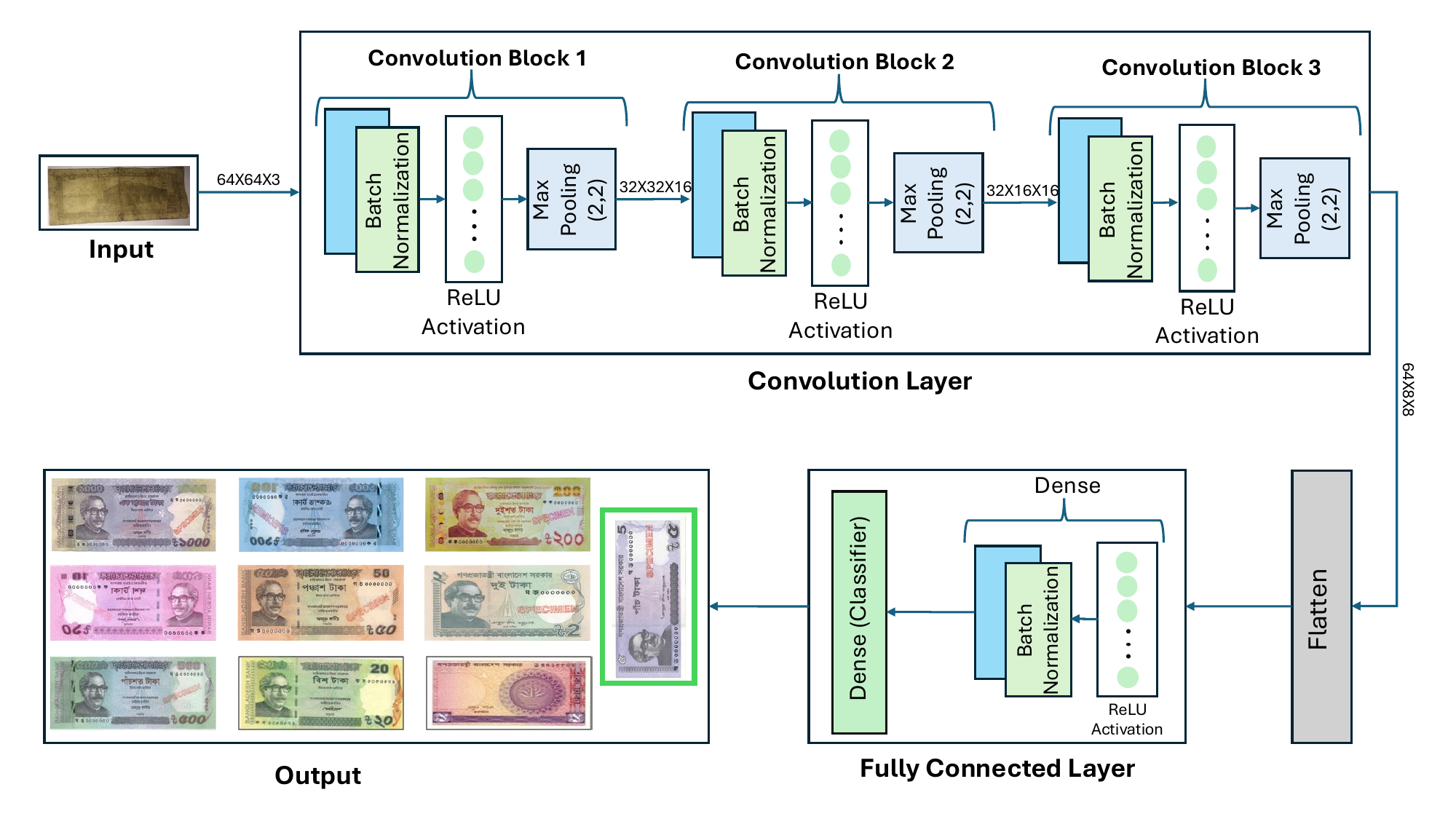}
   \caption{Lightweight CNN Architecture for Currency Classification}
   \label{fig:CustomCNN}
\end{figure}

\textbf{2. Custom CNN Architecture.} We designed a compact CNN (Figure~\ref{fig:CustomCNN}) with three convolutional blocks, each consisting of convolution, batch normalization, ReLU activation, and max pooling. It processes $64\times64$ RGB images with increasing filter counts (16, 32, 64), followed by two dense layers and a softmax output. With only 287k parameters, this model offers efficient inference for low-resource settings.

\textbf{3. Hyperparameter Tuning.}  
Grid search was used to tune learning rate ($1 \times 10^{-5}$ to $5 \times 10^{-3}$), batch size (16 to 128), and number of training epochs (25 to 50). A fixed training duration of 25 epochs was selected based on convergence behavior and computational cost.

\textbf{4. Evaluation Metrics.}  
We evaluated model performance using accuracy, precision, recall, F1-score, confusion matrices, and AUC-ROC. Loss curves were monitored to verify convergence. Parameter counts and model sizes were also recorded to assess deployment feasibility.

\subsection{Damage Analysis Framework}

This work introduces a novel framework for quantifying physical degradation in currency notes through structured comparison with clean reference images. The framework operates along three axes-structural, chromatic, and feature-level integrity-each producing an independent damage metric contributing to a unified usability assessment.

\textbf{1. Background Removal.}  
The input image is converted to HSV color space, and a binary mask is extracted using a threshold $\tau$ on the saturation channel. Morphological filtering removes small artifacts:
\begin{equation}
\text{mask}(x, y) =
\begin{cases}
1, & \text{if } \text{Sat}(x, y) > \tau \\
0, & \text{otherwise}
\end{cases}
\end{equation}
Masked background pixels are replaced with white to isolate the currency region.

\textbf{2. Image Alignment.}  
SIFT keypoints are extracted from both input and reference images, and matched using a FLANN-based approach. RANSAC is applied to estimate a homography matrix $H$:
\begin{equation}
\hat{I}_{\text{damage}}(x, y) = I_{\text{damage}}(H^{-1}(x, y))
\end{equation}

\textbf{3. Binary Damage Estimation.}  
Foreground masks $M_{\text{ref}}$ and $M_{\text{damage}}$ are used to identify missing regions. Binary damage is computed as:
\begin{equation}
\text{BinaryDamage}(\%) = 
\frac{\sum (M_{\text{ref}} - M_{\text{damage}})}{\sum M_{\text{ref}}} \times 100
\end{equation}

\textbf{4. RGB Damage Metric.}  
CLAHE and bilateral filtering enhance image quality before computing the chromatic deviation over the foreground:
\begin{equation}
\begin{aligned}
\text{RGBDamage}(\%) = \frac{100}{255N} \sum_{(x,y)} 
\left| \hat{I}_{\text{damage}}(x,y) - I_{\text{ref}}(x,y) \right| \\
\cdot \mathbf{1}_{\text{mask}}(x,y)
\end{aligned}
\end{equation}
where $N$ is the number of valid pixels.

\textbf{5. Structural Loss in Key Regions.}  
Reference masks $R_k$ define edges and corners. Damage overlap is evaluated as:
\begin{equation}
\text{Overlap}_k = 
\frac{\sum D_{\text{binary}} \cdot R_k}{\sum R_k}
\end{equation}
Regions with $\text{Overlap}_k > 0.1$ are considered structurally damaged.

\begin{algorithm}[ht]
\SetAlgoLined
\SetKwInOut{Input}{Input}
\SetKwInOut{Output}{Output}
\Input{
    $B$: Binary damage percentage, \\
    $R$: RGB damage percentage, \\
    $E$: Number of damaged edges (max 4), \\
    $C$: Number of damaged corners (max 4), \\
    $F$: Count of missing key features, \\
    $N$: Total number of detected features, \\
    $Z$: Number of distinct damaged regions
}
\Output{Final score $\texttt{UCDI} \in [0, 1]$}

$\epsilon \gets 1e{-}5$ \tcp*{Small constant to prevent division by zero}
$Z_{max} \gets 20$ \tcp*{Empirically chosen max damage clusters}

\tcp{Step 1: Normalize inputs}
$B' \gets B / (100 + \epsilon)$\;
$R' \gets R / (100 + \epsilon)$\;
$E' \gets E / 4$\;
$C' \gets C / 4$\;
$F' \gets F / (N + \epsilon)$\;
$Z' \gets Z / (Z_{max} + \epsilon)$\;

\tcp{Step 2: Nonlinear transformations}
$B_t \gets 2.0 \cdot B'$\;
$R_t \gets 2.0 \cdot R'$ \;
$E_t \gets 1.5 \cdot E'$ \;
$C_t \gets (C')^{1.5}$ \;
$F_t \gets \log(1 + F')$ \;
$Z_t \gets \tanh(Z')$ \;

\tcp{Step 3: Weighted sum of components}
$w \gets [0.4, 0.2, 0.15, 0.15, 0.05, 0.05]$ \;
$\texttt{UCDI} \gets 1 - (w_1 B_t + w_2 R_t + w_3 E_t + w_4 C_t + w_5 F_t + w_6 Z_t)$ \;

\tcp{Step 4: Rule-based override for symbolic degradation}
\If{$F/N > 0.45$ \textbf{and} $B > 5$}{
    $\texttt{UCDI} \gets \min(\texttt{UCDI}, 0.65)$
}

\Return{$\texttt{UCDI}$}
\caption{Unified Currency Damage Index (UCDI) Computation}
\label{alg:ucdi}
\end{algorithm}

\textbf{6. Feature Loss via Template Matching.}  
Salient contours in the reference are clustered using DBSCAN. Each cluster $C_i$ is matched in the warped image using normalized cross-correlation:
\begin{equation}
\text{Score}_i = \max_{(x,y)} 
\left( \text{NCC}(C_i, \hat{I}_{\text{damage}}) \right)
\end{equation}
A region is flagged as missing if $\text{Score}_i < 0.5$.

\textbf{7. Unified Currency Damage Index (UCDI).} To consolidate diverse degradation signals into a single usability score, we introduce the \textbf{Unified Currency Damage Index (UCDI)}. This index ranges from 0 (unusable) to 1 (pristine) and combines normalized estimates from binary damage ($B$), RGB degradation ($R$), edge damage ($E$), corner damage ($C$), symbolic feature loss ($F/N$), and damage clustering ($Z$). 

The formulation applies perceptual weighting to reflect practical importance. Binary mask loss receives the highest weight due to its direct link with torn or missing regions. RGB deviation accounts for wear and staining. Edge and corner terms are shaped to highlight structural instability, with corner degradation exponentiated to emphasize critical areas. Feature loss is scaled logarithmically to limit the penalty for partial symbol degradation. Clustering is regularized using $\tanh$ to contain extreme outliers from spatially concentrated noise.

A conservative cap is applied when symbolic feature loss exceeds 45\% and binary damage is above 5\%, limiting the UCDI to 0.65. This constraint prevents false optimism when structural metrics underestimate real-world usability. The resulting UCDI provides an interpretable and numerically stable representation of physical condition across a wide range of damage types.

\subsubsection*{\textbf{Complexity Analysis of UCDI}}

The overall time complexity of the damage analysis framework is $O(n \log n + k \log k + cn)$, where $n$ is the number of image pixels, $k$ is the number of SIFT keypoints, and $c$ is the number of feature clusters. Most components, including background removal and damage metrics, run in $O(n)$. SIFT-based alignment adds $O(k \log k)$, and DBSCAN-based template matching contributes $O(cn)$, though $c$ is small in practice. Space complexity is $O(n + k)$ due to pixel masks and keypoint descriptors. UCDI weights and thresholds were selected through visual inspection and empirical tuning without supervised damage labels.

\subsection{Counterfeit Detection Framework}

Building on our classification pipeline, we developed a dedicated counterfeit detection module using a dataset specifically designed for this task. Existing classification datasets lacked labeled counterfeit samples and did not reflect the visual characteristics necessary to train robust forgery detectors. To address this, we adopted the NoteShieldBD~\cite{NoteShieldBD} dataset, which includes labeled images of both genuine and counterfeit Bangladeshi banknotes in the 500 and 1000 taka denominations. Each note is photographed from six viewpoints, increasing robustness to orientation and lighting changes. Counterfeit samples were sourced in coordination with law enforcement agencies, ensuring realistic and representative coverage.

The same preprocessing pipeline used in our denomination classification system was applied here. This included noise reduction, contrast enhancement, and local histogram equalization, which helped emphasize subtle differences between authentic and forged notes. These enhancements proved effective in revealing minor structural inconsistencies, color irregularities, and print artifacts typically present in counterfeit specimens.

We evaluated five models for binary classification: VGG16, ResNet50, InceptionV3, EfficientNetB0, and a lightweight custom CNN. For each architecture, the final layers were replaced with binary classification heads, and the upper layers were fine-tuned on the NoteShieldBD dataset. All models were trained for 25 epochs using an 80:10:10 stratified split, with cross-entropy loss as the objective.

\begin{table*}[htbp]
\caption{Performance of Different Models on Individual Datasets: Optimal hyperparameters (batch size, learning rate) and F1-scores across three datasets with varying class distributions.}
\centering
\begin{tabular}{|l|c|cccc|cccc|cccc|}
\hline
\multirow{2}{*}{\textbf{Model}} & \textbf{Params} & \multicolumn{4}{|c|}{\textbf{Dataset 1}} & \multicolumn{4}{|c|}{\textbf{Dataset 2}} & \multicolumn{4}{|c|}{\textbf{Dataset 3}} \\ 
\cline{3-14}
 & & \textbf{Batch} & \textbf{LR} & \textbf{Classes} & \textbf{F1} & \textbf{Batch} & \textbf{LR} & \textbf{Classes} & \textbf{F1} & \textbf{Batch} & \textbf{LR} & \textbf{Classes} & \textbf{F1} \\ 
\hline
\textbf{VGG16}            & ~138M  & 128  & 5e-05  & 9   & 0.9425  & 64  & 0.0001  & 10  & 0.9326  & 128  & 1e-05  & 8   & 0.9187  \\ 
\textbf{ResNet50}         & ~25.6M & 64   & 5e-05  & 9   & 0.9077  & 64  & 0.0001  & 10  & 0.9299  & 64   & 1e-05  & 8   & 0.9221  \\ 
\textbf{EfficientNet\_B0} & ~5.3M  & 64   & 5e-05  & 9   & 0.9034  & 64  & 0.0001  & 10  & \textbf{0.9382}  & 64   & 1e-05  & 8   & 0.8659  \\ 
\textbf{Inception\_V3}    & ~23.8M & 128  & 0.0005 & 9   & \textbf{0.9499}  & 128 & 0.0005  & 10  & 0.9177  & 128  & 5e-05  & 8   & 0.9342  \\ 
\textbf{Custom\_CNN}      & ~287k  & 128  & 0.0001 & 9   & 0.9238  & 128 & 0.0001  & 10  & 0.8657  & 128  & 5e-05  & 8   & \textbf{0.9581}  \\ 
\hline
\end{tabular}
\label{tab:model_performance}
\end{table*}

\begin{table*}[htbp]
\caption{Performance Comparison on Combined Dataset: Evaluation of all models using consistent training configuration on the unified 10-class dataset, reporting F1-score and model file size for deployment assessment.}
\centering
\begin{tabular}{|l|c|c|c|c|c|c|c|}
\hline
\textbf{Model}          & \textbf{Params} & \textbf{Batch} & \textbf{LR}    & \textbf{Epochs} & \textbf{Classes} & \textbf{F1}   & \textbf{Size (MB)} \\ 
\hline
\textbf{VGG16}          & ~138M           & 64             & 1e-05          & 25              & 10               & 0.9315        & 512               \\ 
\textbf{ResNet50}       & ~25.6M         & 128            & 5e-05          & 25              & 10               & 0.9176        & 90.1              \\ 
\textbf{EfficientNet\_B0} & ~5.3M          & 128            & 5e-05          & 25              & 10               & 0.9417        & 15.6              \\ 
\textbf{Inception\_V3}  & ~23.8M         & 64             & 0.0001         & 25              & 10               & 0.9519        & 83.5              \\ 
\textbf{Custom\_CNN}    & \textbf{~287k}          & 128            & 0.0001         & 25              & 10               & 0.9342        & \textbf{1.1}               \\ 
\hline
\end{tabular}
\label{tab:combined_performance}
\end{table*}

Accurate and low-latency counterfeit detection is essential for real-world viability, especially in mobile or embedded environments and for users relying on assistive technologies. To meet these requirements, we evaluated not only classification accuracy but also model size and inference speed. The goal is to ensure a secure, responsive, and deployable system without compromising detection quality.


\section{Experimental Setup}

All experiments were conducted using Python 3.9.9 and PyTorch 1.12.1 with CUDA 11.3 on an NVIDIA A100-SXM4-80GB GPU, with up to 40GB memory allocated per run. Numerical operations were supported by NumPy 1.21.4, and image preprocessing was handled using OpenCV and PIL. The entire pipeline was divided into three modules: currency classification, damage analysis, and counterfeit detection.

To support reproducibility, we release the complete codebase, including training and evaluation scripts, configuration files, and pretrained model weights. All software dependencies are specified in the repository.

\vspace{1em}
\noindent\textbf{Code Repository:} \href{https://github.com/AnonymousResearcher-ai/Currency-Evaluation}{GitHub Link}

\begin{figure*}[htbp]
    \centering
    \begin{subfigure}[t]{0.3\textwidth}
        \centering
        \includegraphics[width=\textwidth]{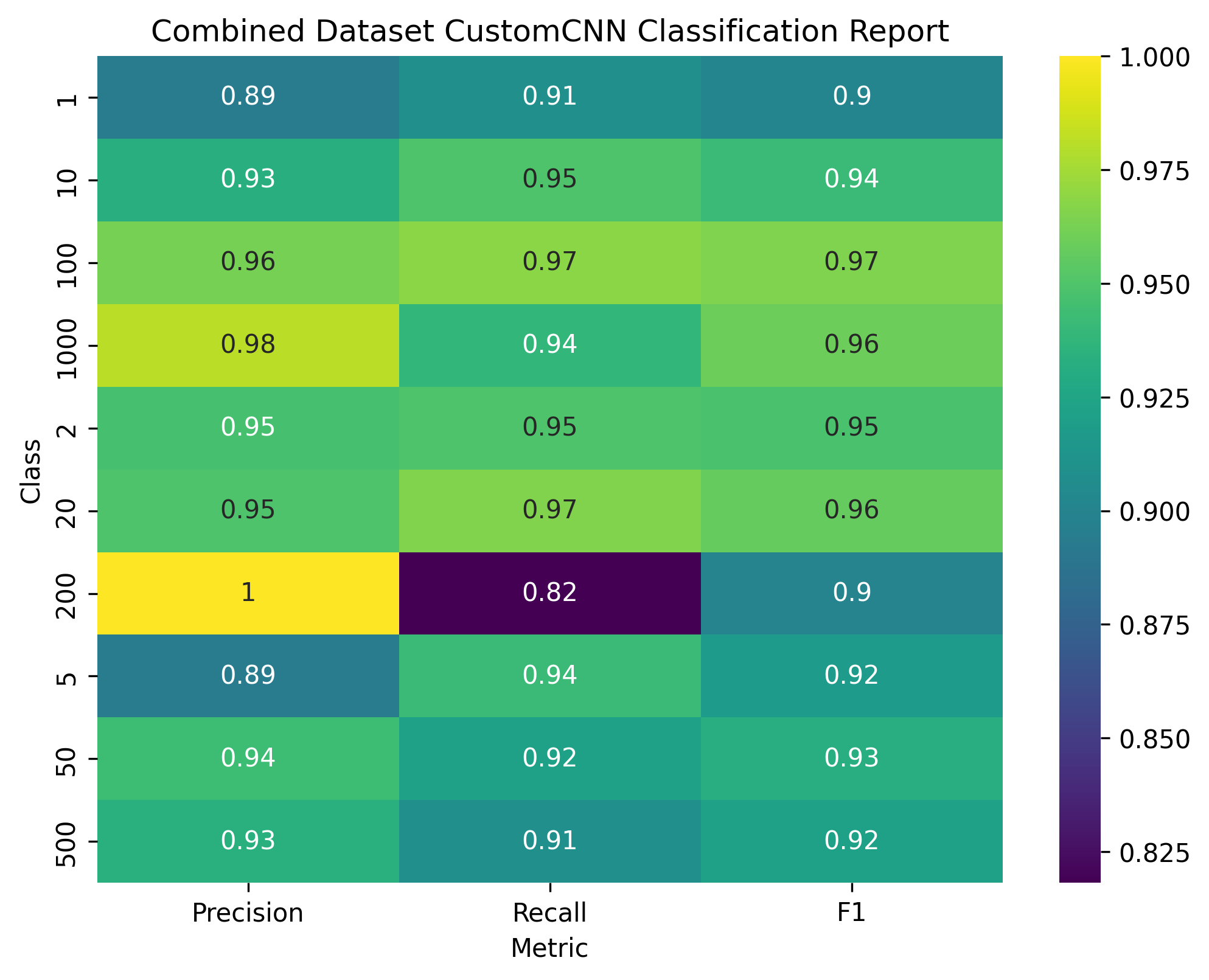}
        \caption{Per-class precision, recall, and F1 scores for Custom\_CNN.}
        \label{fig:custom_cnn_cr}
    \end{subfigure}
    \hfill
    \begin{subfigure}[t]{0.3\textwidth}
        \centering
        \includegraphics[width=\textwidth]{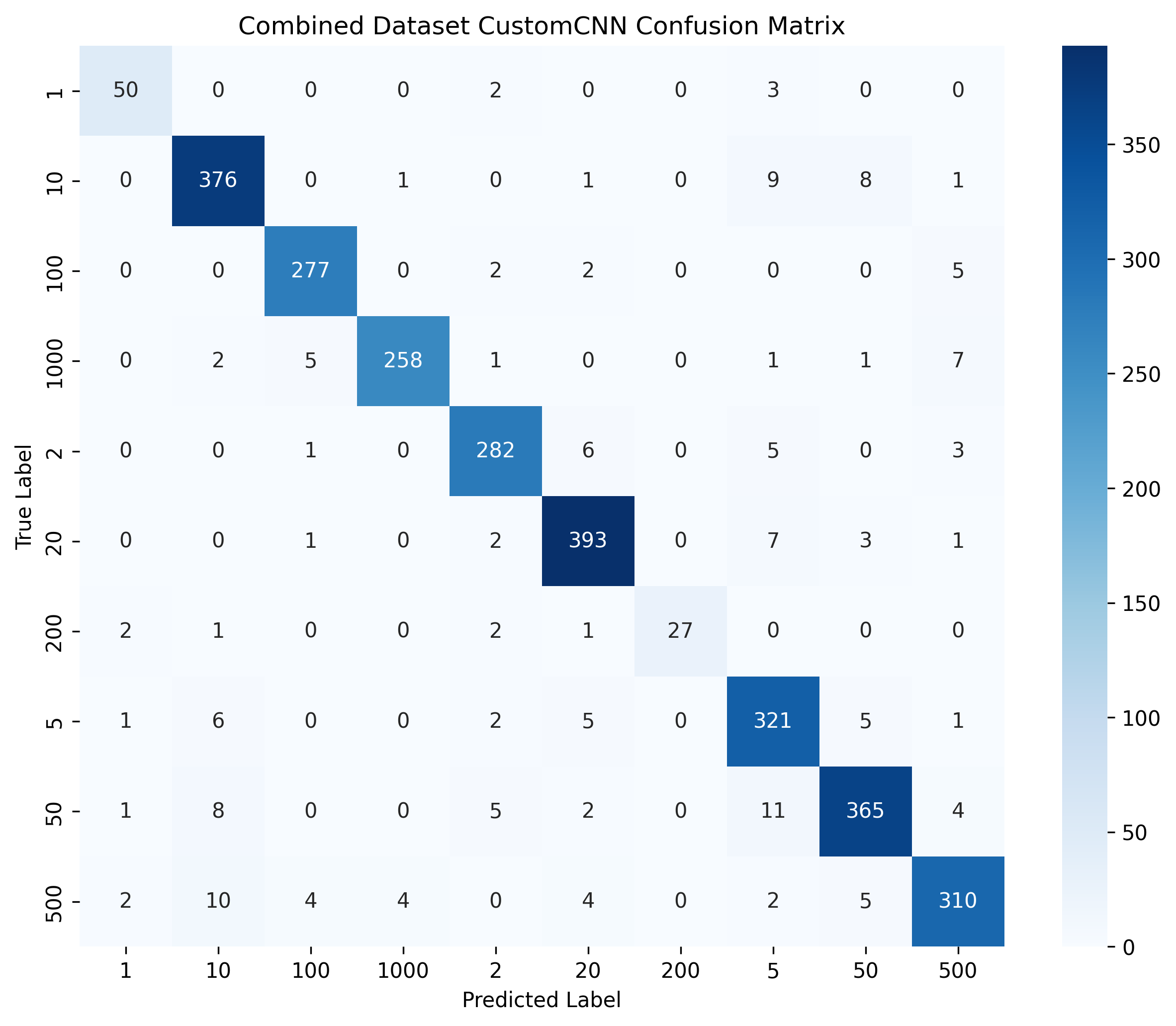}
        \caption{Confusion matrix showing class-wise prediction distribution.}
        \label{fig:custom_cnn_confusion_matrix}
    \end{subfigure}
    \hfill
    \begin{subfigure}[t]{0.3\textwidth}
        \centering
        \includegraphics[width=\textwidth]{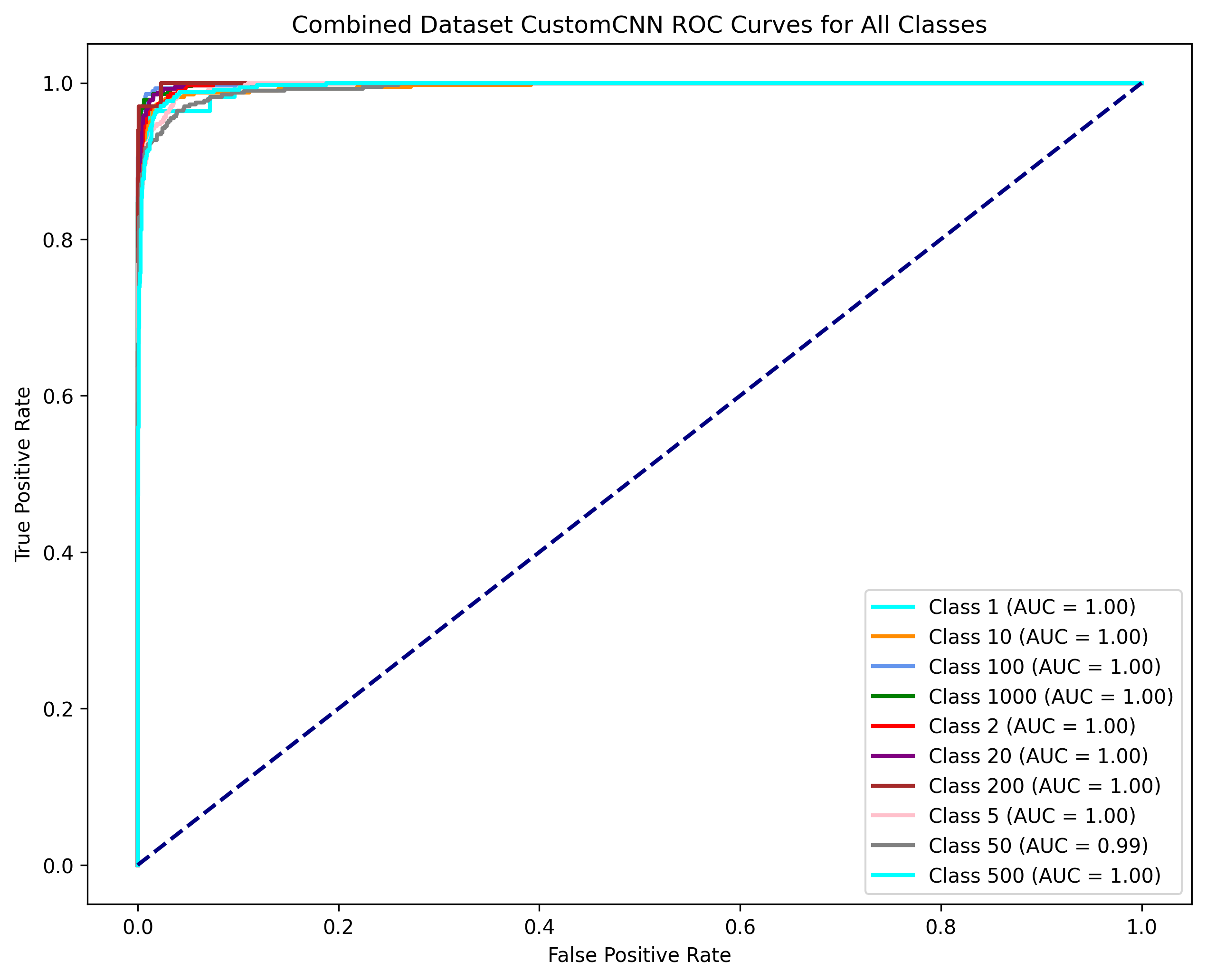}
        \caption{ROC curves across all currency classes for Custom\_CNN.}
        \label{fig:custom_cnn_roc}
    \end{subfigure}
    \vskip\baselineskip
    \hspace{\fill}
    \begin{subfigure}[t]{0.3\textwidth}
        \centering
        \includegraphics[width=\textwidth]{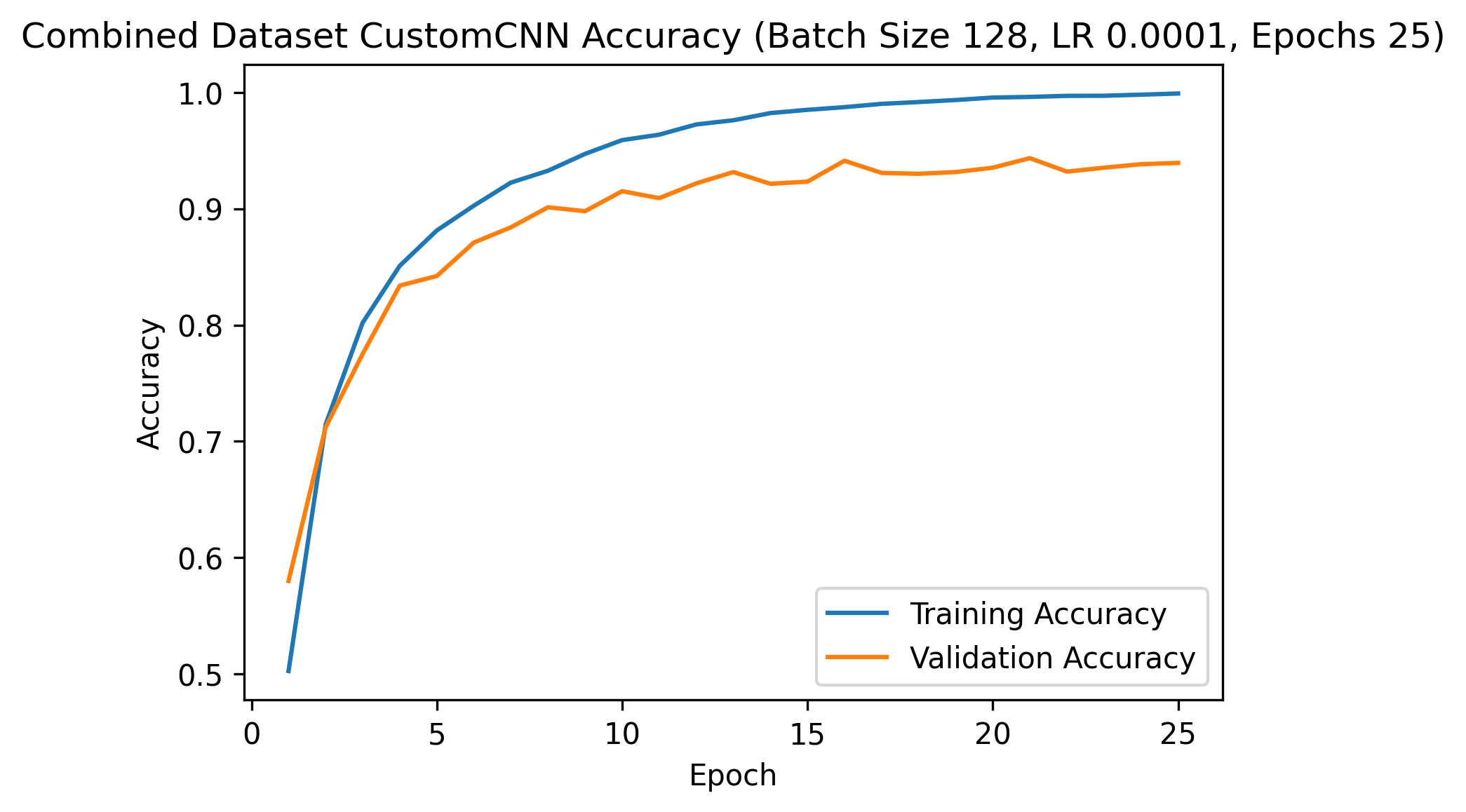}
        \caption{Training and validation accuracy over epochs.}
        \label{fig:custom_cnn_acc}
    \end{subfigure}
    \hspace{0.1\textwidth} 
    \begin{subfigure}[t]{0.3\textwidth}
        \centering
        \includegraphics[width=\textwidth]{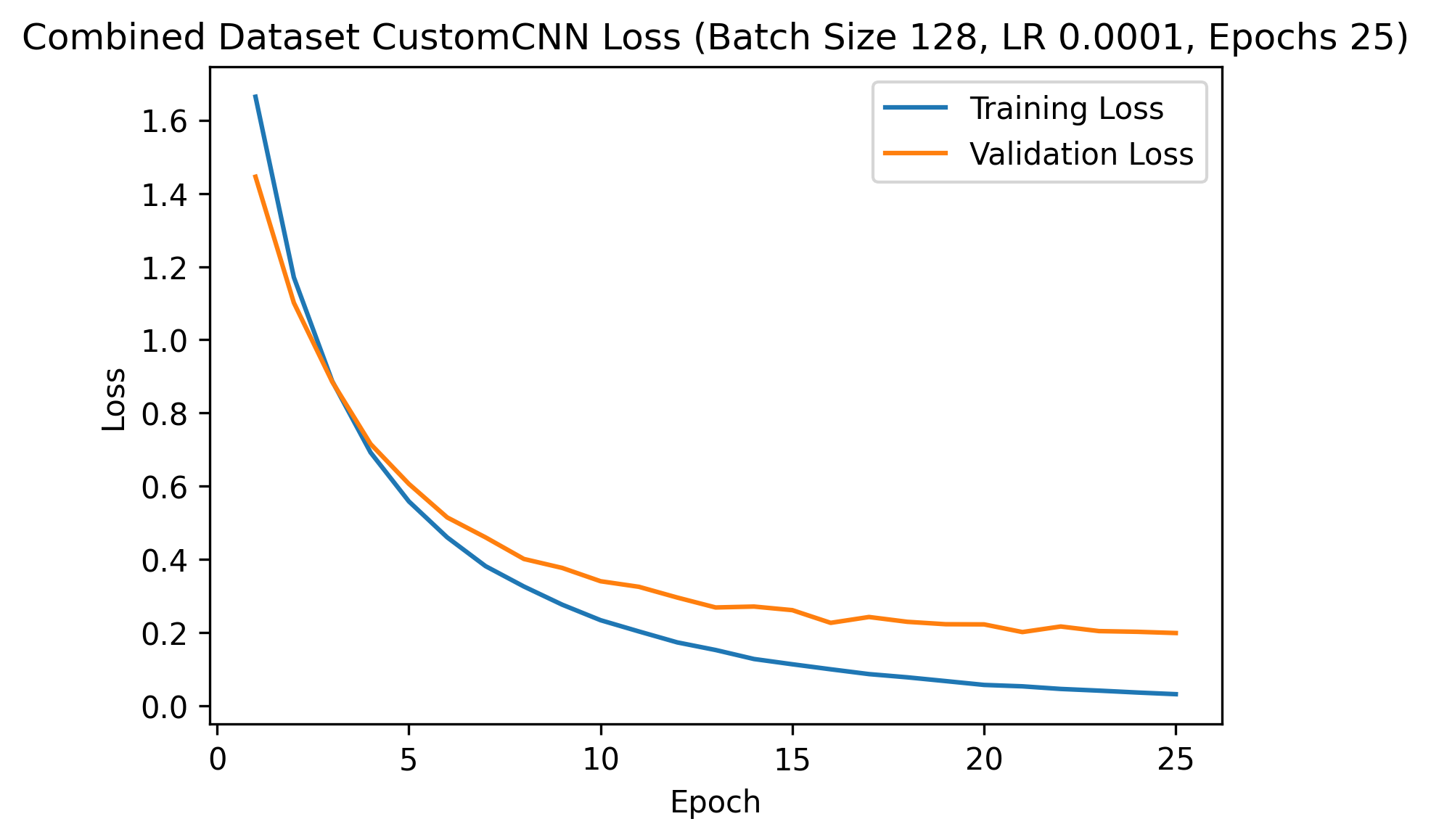}
        \caption{Training and validation loss convergence over epochs.}
        \label{fig:custom_cnn_loss}
    \end{subfigure}
    \hspace{\fill}

    \caption{Evaluation of Custom\_CNN on the combined dataset, showing per-class metrics, confusion matrix, ROC curves, and training performance over epochs.}
    \label{fig:classification_reports}
\end{figure*}


\section{Results and Discussion}

\subsection*{\textbf{Currency Classification Results}}

We evaluated five convolutional neural networks on three individual datasets and one combined dataset to assess the accuracy and efficiency of our currency classification framework. Tables~\ref{tab:model_performance} and~\ref{tab:combined_performance} present comparative performance metrics, while Figure~\ref{fig:classification_reports} provides a breakdown of Custom\_CNN's diagnostics.

Table~\ref{tab:model_performance} compares model performance across datasets with distinct noise levels and image quality. On Dataset~1, InceptionV3 achieved the highest F1-score of 0.9499, demonstrating robustness to noisy and low-resolution inputs. EfficientNetB0 performed best on Dataset~2 (F1-score 0.9382), which features balanced, high-resolution samples. Dataset~3 posed more intra-class variability, yet our Custom\_CNN achieved an F1-score of 0.9581, outperforming larger models despite having only 287K parameters. This result shows that lightweight models can generalize well in diverse settings when properly tuned. Model performance varied with data characteristics. Larger networks benefitted from clean, high-quality inputs, while compact architectures like Custom\_CNN performed strongly on real-world, heterogeneous datasets.

To simulate deployment conditions, we merged all datasets into a 10-class corpus. InceptionV3 led with an F1-score of 0.9519, followed by EfficientNetB0 (0.9417) and Custom\_CNN (0.9342). Despite its size (83MB), InceptionV3 only marginally outperformed Custom\_CNN, which is 1.1MB and highly suitable for resource-constrained devices. This accuracy-efficiency tradeoff is critical in real-world deployments. Custom\_CNN achieves near-state-of-the-art results while remaining deployable on mobile or embedded systems.

Figure~\ref{fig:classification_reports} illustrates the diagnostic evaluation of Custom\_CNN on the combined dataset. The classification report indicates high precision and recall across most currency classes, with F1-scores exceeding 0.90 for commonly used denominations. Slight drops were observed in lower-value classes, such as the 1-unit note, where intra-class variance and limited data volume can affect feature generalization. 

The confusion matrix reveals strong class-wise separation for dominant categories such as 100, 500, and 1000 units. Some confusion was observed between visually similar notes (e.g., 5 and 50 units), a pattern consistent across all evaluated models. These results suggest that while overall classification is reliable, incorporating additional discriminative features (e.g., serial numbers or symbols) may further reduce misclassifications in future iterations.

The ROC curve shows sharp separation between true and false positives across all classes, validating the model’s ability to maintain decision boundaries in a multi-class setting. The training accuracy and loss plots further confirm stable convergence under the selected hyperparameters, with no signs of overfitting or instability.

Table~\ref{tab:combined_performance} also provides model sizes, a key factor in real-world deployment. VGG16, while historically effective, requires over 500MB of storage and is impractical for constrained environments. In contrast, EfficientNetB0 offers a favorable balance with high accuracy and reduced size (15.6MB), making it suitable for most modern smartphones. However, Custom\_CNN achieves the smallest footprint (1.1MB) while maintaining performance competitive with the state-of-the-art. With a batch size of 128 and minimal memory requirements, Custom\_CNN enables real-time inference on entry-level hardware, supporting deployment in mobile and embedded systems. This is especially valuable for use cases involving offline processing, intermittent connectivity, or decentralized validation workflows.

Unlike most prior studies which focus on single datasets or constrained conditions, our evaluation spans three distinct public datasets, collectively representing varied imaging scenarios. Compared to Pach\'on et al.~\cite{pachon}, who demonstrated compact CNNs on Colombian currency, and Jaman et al.~\cite{jaman2025bd}, who emphasized offline utility, our Custom\_CNN achieves comparable or better performance with significantly less memory usage across a more diverse image pool. Tasnim et al.~\cite{tasnim2021bangladeshi} and Das et al.~\cite{das2023utilizing} prioritized accessibility, but did not benchmark across heterogeneous datasets. Our results show that the Custom\_CNN offers reliable accuracy in conditions reflecting practical deployment, reinforcing its suitability for low-resource, real-world applications.

\subsection*{\textbf{Damage Detection Results}}

\begin{figure}[htbp]
    \centering
    \begin{subfigure}[b]{0.45\linewidth}
        \includegraphics[width=\linewidth]{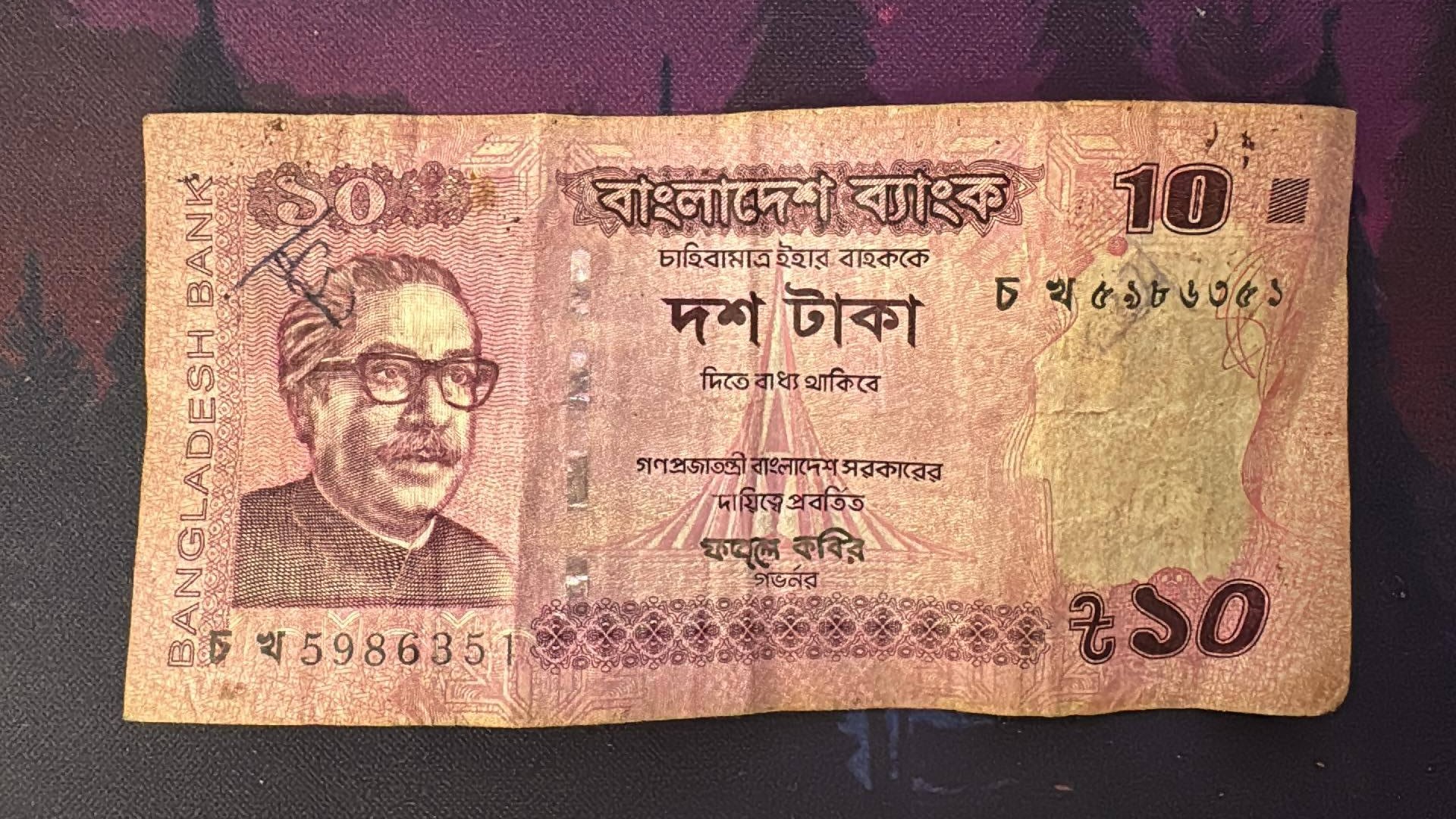}
        \caption{Case 1}
    \end{subfigure}
    \hfill
    \begin{subfigure}[b]{0.45\linewidth}
        \includegraphics[width=\linewidth]{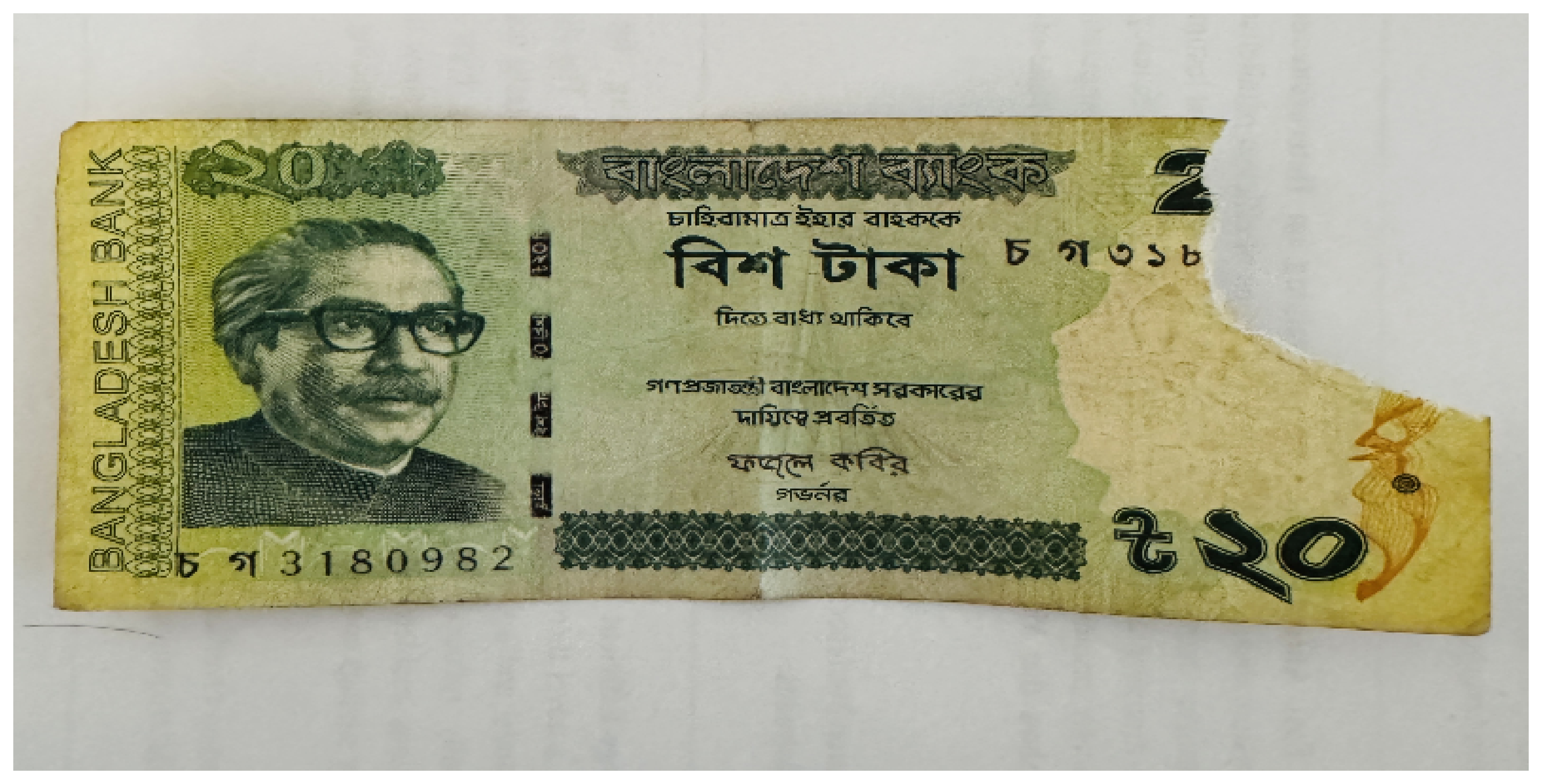}
        \caption{Case 2}
    \end{subfigure}
    \vspace{1mm}
    \begin{subfigure}[b]{0.45\linewidth}
        \includegraphics[width=\linewidth]{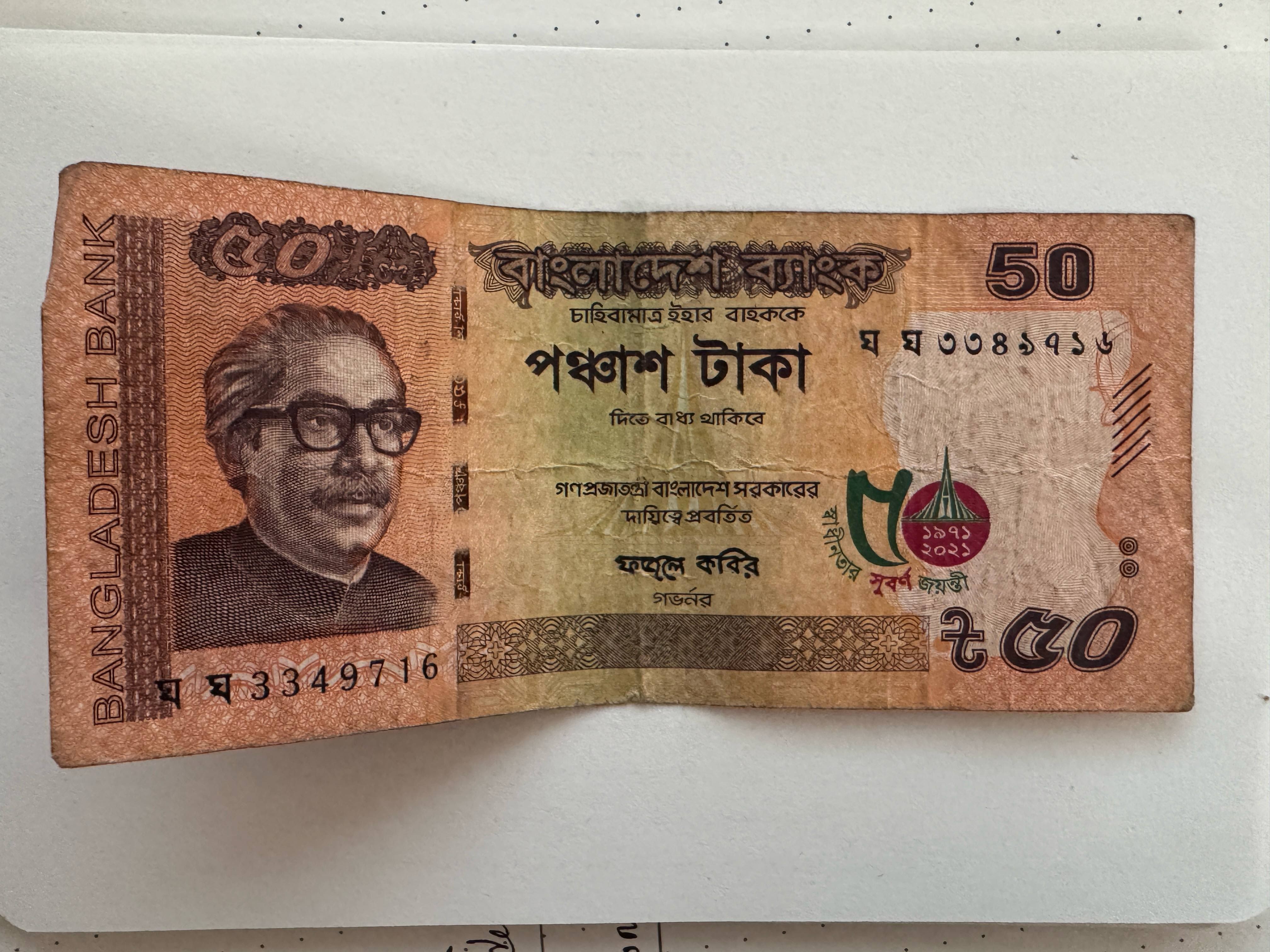}
        \caption{Case 3}
    \end{subfigure}
    \hfill
    \begin{subfigure}[b]{0.45\linewidth}
        \includegraphics[width=\linewidth]{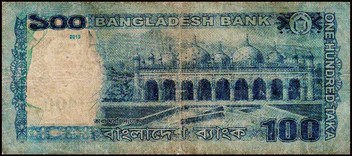}
        \caption{Case 4}
    \end{subfigure}
    \caption{Sample images of physically damaged currency notes used for usability assessment. Each case corresponds to a row in Table~\ref{tab:ucdi_summary}.}
    \label{fig:torn_notes}
\end{figure}

\begin{figure}[htbp]
    \centering
    \includegraphics[width=0.45\linewidth]{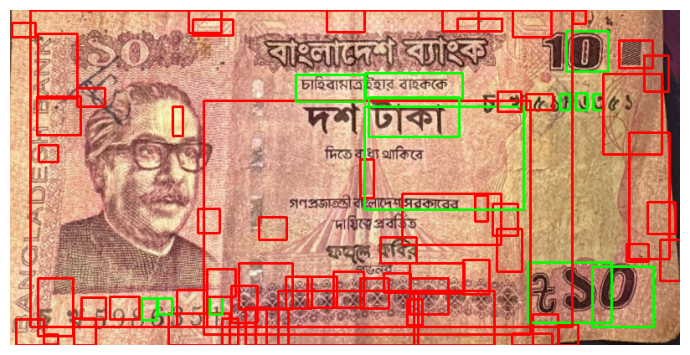}
    \includegraphics[width=0.45\linewidth]{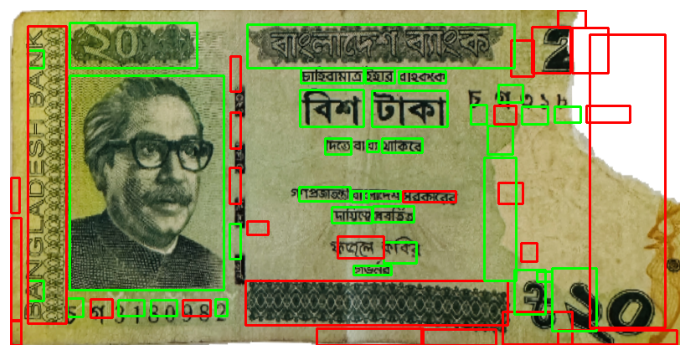}
    \vspace{1mm}
    \includegraphics[width=0.45\linewidth]{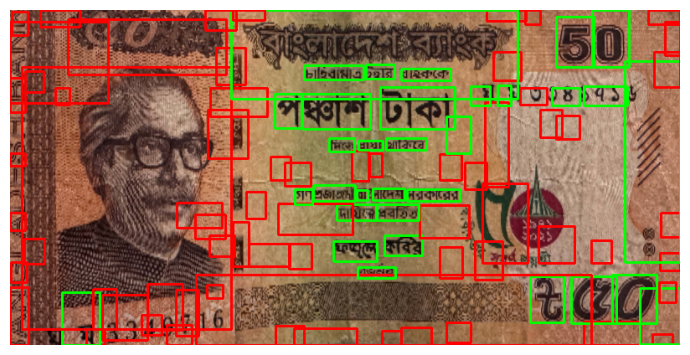}
    \includegraphics[width=0.45\linewidth]{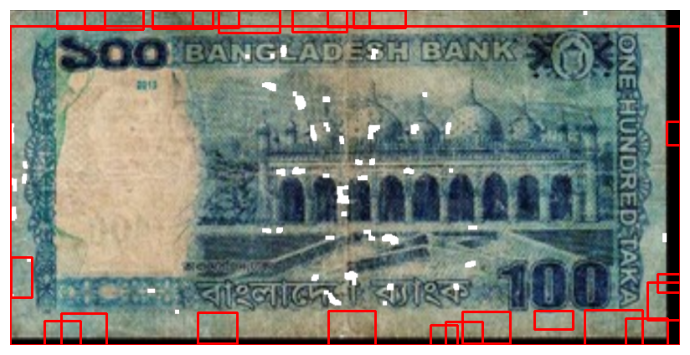}
    \caption{Detected features and contours from damaged currency notes using DBSCAN clustering.}
    \label{fig:feature_detection}
\end{figure}

\begin{figure}[htbp]
    \centering
    \includegraphics[width=0.45\linewidth]{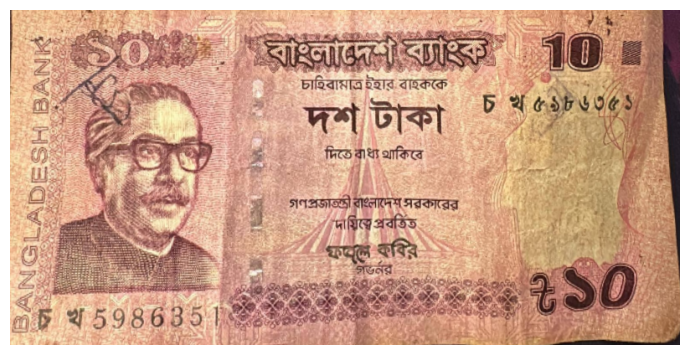}
    \includegraphics[width=0.45\linewidth]{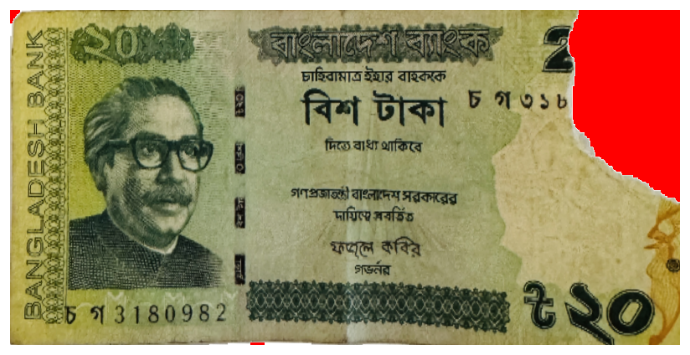}
    \vspace{1mm}
    \includegraphics[width=0.45\linewidth]{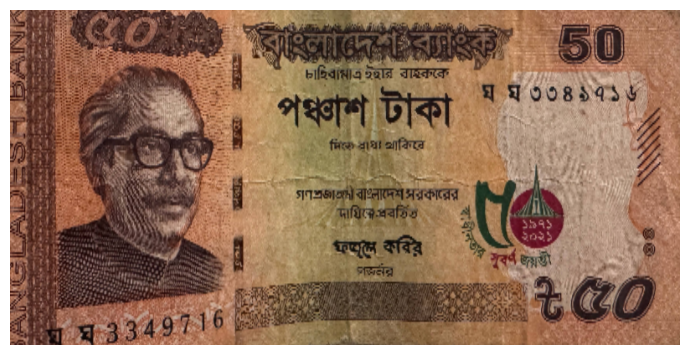}
    \includegraphics[width=0.45\linewidth]{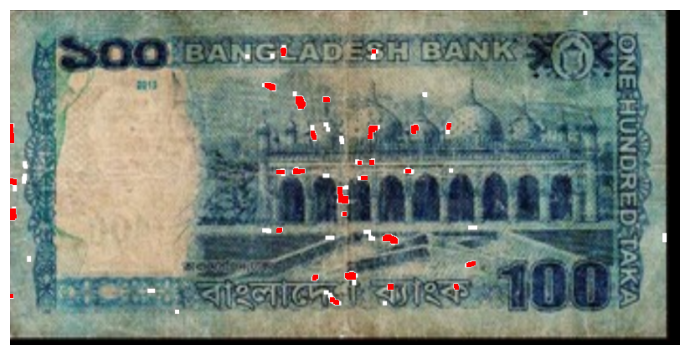}
    \caption{Binary mask-based torn area estimation compared to reference notes.}
    \label{fig:binary_damage}
\end{figure}

\begin{figure}[htbp]
    \centering
    \includegraphics[width=0.45\linewidth]{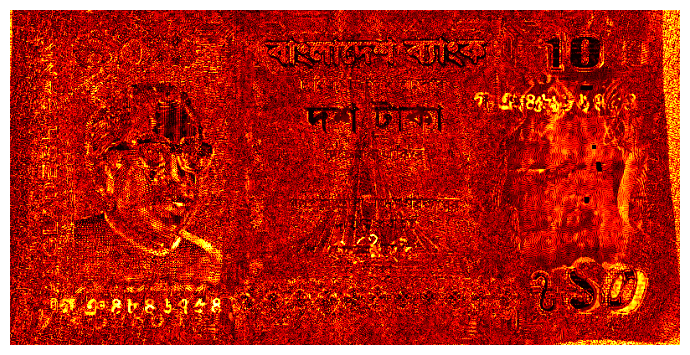}
    \includegraphics[width=0.45\linewidth]{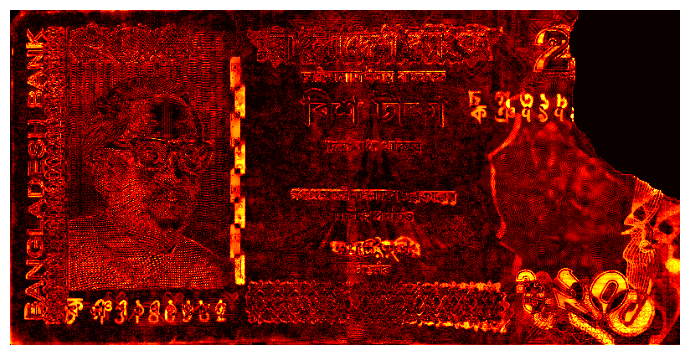}
    \vspace{1mm}
    \includegraphics[width=0.45\linewidth]{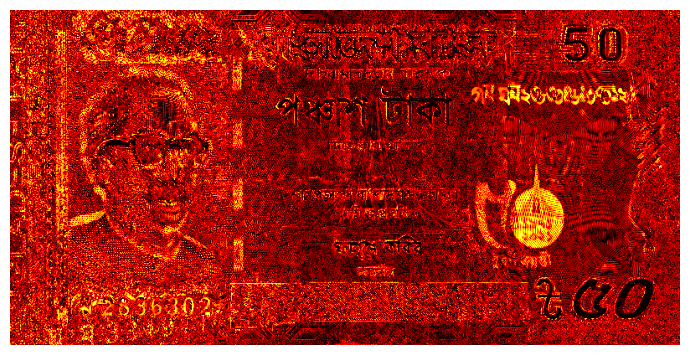}
    \includegraphics[width=0.45\linewidth]{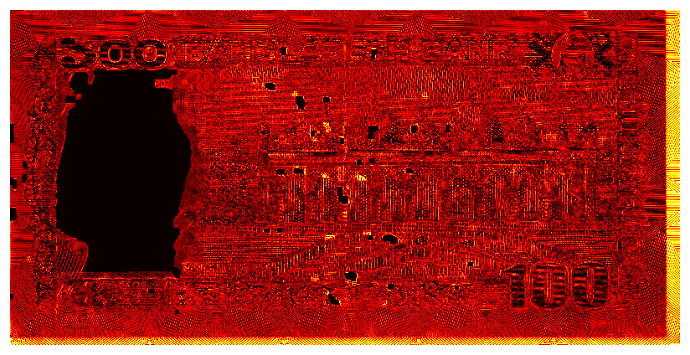}
    \caption{RGB pixel-wise difference analysis between damaged and reference currency notes.}
    \label{fig:rgb_damage}
\end{figure}

We evaluated four real-world currency notes to demonstrate the effectiveness of our damage quantification framework. Each case was analyzed using the proposed Unified Currency Damage Index (UCDI), which consolidates structural, chromatic, and symbolic degradation into a single usability score. Table~\ref{tab:ucdi_summary} presents the detailed measurements across damage axes, while Figure~\ref{fig:torn_notes} shows the corresponding currency samples. The intermediate outputs in Figures~\ref{fig:feature_detection}, \ref{fig:binary_damage}, and \ref{fig:rgb_damage} illustrate feature extraction, binary mask alignment, and RGB deviation used in score computation.

\textbf{Case 1} shows a visually faded note with no physical tears or missing features in structure, but substantial chromatic distortion (24.13\% RGB damage) and 60 out of 73 key features missing. Despite zero binary damage and no structural failure, the loss of symbolic regions lowers its UCDI to 0.8735, indicating moderate usability with partial visual degradation.

\textbf{Case 2} presents a more visibly compromised note, exhibiting localized tears and missing corner content. With 7.69\% binary damage, 3 damaged structural regions, and 26 out of 60 features missing, the model detects significant degradation both structurally and semantically. The UCDI score of 0.7264 places it near the lower operational limit, reflecting a high probability of rejection in automated systems.

\textbf{Case 3}, although slightly faded (21.0\% RGB damage), maintains structural integrity and binary completeness. The note shows no damaged edges or corners, and 67 out of 100 key features were lost, largely due to uniform wear and low-contrast regions. The final UCDI of 0.8905 supports the interpretation that despite visual wear, the note remains highly usable.

\begin{table*}[htbp]
\caption{Damage Analysis Summary for Sample Notes: Evaluation of structural and visual degradation using binary masks, RGB deviation, and feature loss. UCDI reflects overall usability.}
\centering
\begin{tabular}{|c|c|c|c|c|c|c|}
\hline
\textbf{Sample} & \textbf{Binary Damage (\%)} & \textbf{RGB Damage (\%)} & \textbf{Damaged Edges} & \textbf{Damaged Corners} & \textbf{Missing / Total Features} & \textbf{UCDI Score} \\
\hline
Case 1 & 0.0000 & 24.1346 & 0 & 0 & 60 / 73  & 0.8735 \\
Case 2 & 7.6906 & 13.8539 & 2 & 1 & 26 / 60  & 0.7264 \\
Case 3 & 0.0000 & 20.9730 & 0 & 0 & 67 / 100 & 0.8905 \\
Case 4 & 0.6890 & 22.6983 & 0 & 0 & 23 / 23  & 0.8373 \\
\hline
\end{tabular}
\label{tab:ucdi_summary}
\end{table*}

\textbf{Case 4} demonstrates minimal physical tearing (0.69\% binary damage) and no edge or corner failures. However, all 23 detected symbolic features were classified as missing, due to significant blur and contrast loss in the visual field. Despite this, the low structural damage and RGB deviation (22.69\%) produce a UCDI of 0.8373. This case highlights how symbolic feature absence alone can drive usability scores downward, even when the note remains structurally intact.

Overall, the results validate the robustness and sensitivity of our UCDI-based framework. The score distribution across the four cases remains in the range of 0.72-0.89, avoiding extreme polarity and offering nuanced interpretation. This continuous scoring method offers a practical alternative to binary classification for deployment in automated currency validation systems.

\subsection*{\textbf{Counterfeit Detection Results}}

Table~\ref{tab:fake-currency} summarizes the classification performance across five CNN models. ResNet50 and the Custom CNN achieved the highest accuracy (0.99), with strong and balanced precision, recall, and F1-scores for both genuine and counterfeit classes. InceptionV3 followed closely, while EfficientNet-B0 and VGG16 trailed slightly, the latter showing reduced recall (0.90) for counterfeit notes.

The Custom CNN, despite being only 1.1 MB in size, matched ResNet50’s performance while outperforming heavier models like VGG16 (512 MB) and InceptionV3 (83.5 MB) in efficiency. With an F1-score of 0.9342 and low computational overhead, it offers a practical solution for deployment on mobile and edge devices.

Prior works report accuracy ranging from 0.90 to 0.99 \cite{Bandu, bhatia2021, latha2021}, but often with models unsuitable for real-time or low-power contexts. Our results show that high detection performance can be retained without large-scale architectures, addressing both accuracy and deployability in a unified system.

\begin{table}[htbp]
\centering
\caption{Classification performance for genuine and counterfeit notes on the NoteShieldBD dataset. The Custom CNN achieves top accuracy with minimal model size.}
\begin{tabular}{|cccccc|}
\hline
\multicolumn{1}{|c|}{\textbf{Model}} & \multicolumn{1}{c|}{\textbf{Class}} & \multicolumn{1}{c|}{\textbf{Acc.}} & \multicolumn{1}{c|}{\textbf{Prec.}} & \multicolumn{1}{c|}{\textbf{Recall}} & \textbf{F1} \\ \hline
\multirow{2}{*}{VGG16}               & Genuine                             & \multirow{2}{*}{0.95}              & 0.93                                & 0.99                                 & 0.96        \\
                                     & Counterfeit                         &                                    & 0.99                                & 0.90                                 & 0.95        \\ \hline
\multirow{2}{*}{ResNet50}            & Genuine                             & \multirow{2}{*}{0.99}              & 0.99                                & 0.99                                 & 0.99        \\
                                     & Counterfeit                         &                                    & 0.98                                & 0.98                                 & 0.98        \\ \hline
\multirow{2}{*}{EffNet-B0}           & Genuine                             & \multirow{2}{*}{0.97}              & 0.97                                & 0.98                                 & 0.97        \\
                                     & Counterfeit                         &                                    & 0.97                                & 0.96                                 & 0.96        \\ \hline
\multirow{2}{*}{IncepV3}             & Genuine                             & \multirow{2}{*}{0.98}              & 0.98                                & 0.99                                 & 0.98        \\
                                     & Counterfeit                         &                                    & 0.98                                & 0.98                                 & 0.98        \\ \hline
\multirow{2}{*}{Custom CNN}          & Genuine                             & \multirow{2}{*}{0.99}              & 0.99                                & 0.99                                 & 0.99        \\
                                     & Counterfeit                         &                                    & 0.98                                & 0.98                                 & 0.98        \\ \hline
\end{tabular}
\label{tab:fake-currency}
\end{table}

\section{Conclusion}

This paper introduces a unified framework for currency evaluation tailored to low-resource environments, addressing the limitations of existing systems that handle classification, damage assessment, and counterfeit detection in isolation. The proposed pipeline integrates these components efficiently, with an emphasis on deployment feasibility and real-world applicability. Experimental results demonstrate that our lightweight Custom\_CNN achieves competitive classification performance with minimal computational cost. The proposed \textbf{Unified Currency Damage Index (UCDI)} provides a structured, interpretable metric for quantifying physical degradation, while the counterfeit detection module offers robust discrimination under varied imaging conditions.

These findings show the potential for compact, modular tools that support inclusive currency validation across diverse usage scenarios. The system meets the practical requirements of offline deployment, real-time inference, and adaptability to varying note conditions. Key contributions include a deployable CNN-based classification model, a multi-factor damage scoring framework, and an effective counterfeit detection method. All components are integrated into a single pipeline and optimized for constrained settings.

Future work will focus on validating the damage model using annotated datasets that reflect real-world impairments, including handwritten annotations, taped repairs, and layered degradation. Expanding the framework to handle such complex cases remains a necessary step toward broader adoption in unstructured environments.

\end{document}